# Fractional Moments on Bandit Problems

**Ananda Narayanan B**
Dept. of Electrical Engg.
Indian Institute of Technology Madras
ananda@smail.iitm.ac.in

**Balaraman Ravindran**
Dept. of Computer Science and Engg.
Indian Institute of Technology Madras
ravi@cse.iitm.ac.in

## Abstract

Reinforcement learning addresses the dilemma between exploration to find profitable actions and exploitation to act according to the best observations already made. Bandit problems are one such class of problems in stateless environments that represent this explore/exploit situation. We propose a learning algorithm for bandit problems based on fractional expectation of rewards acquired. The algorithm is theoretically shown to converge on an $\epsilon$-optimal arm and achieve $O(n)$ sample complexity. Experimental results show the algorithm incurs substantially lower regrets than parameter-optimized $\epsilon$-greedy and SoftMax approaches and other low sample complexity state-of-the-art techniques.

## 1 INTRODUCTION

The multi-arm bandit problem captures general aspects of learning in an unknown environment [Berry & Fristedt, 1985]. Decision theoretic issues like how to minimize learning time, and when to stop learning and start exploiting the knowledge acquired are well embraced in the multi-arm bandit problem. This problem is of useful concern in different areas of artificial intelligence such as reinforcement learning [Sutton & Barto, 1998] and evolutionary programming [Holland, 1992]. The problem also has applications in many fields including industrial engineering, simulation and evolutionary computation [Kalyanakrishnan & Stone, 2010].

An n-arm bandit problem is to learn to preferentially select a particular action (or pull a particular arm) from a set of n actions (arms) numbered $1, 2, 3, \ldots, n$. Each selection of an action results in a reward that follows a distribution governing a random variable $R_i$ with mean $\mu_i$ usually derived from a stationary probability distribution corresponding to the action being selected. Arms are pulled and rewards acquired until learning converges. If $\mu^* = max_i\{\mu_i\}$, then arm $j$ is defined to be $\epsilon$-optimal if $\mu_j > \mu^* - \epsilon$. An arm corresponding to a mean equal to $\mu^*$ is considered the best arm as it is expected to give the maximum reward per pull. Let $Z$ be the total reward acquired over $l$ successive pulls. Define regret $\eta$ as the loss in total reward acquired had the best arm been repetitively chosen right from the first pull, i.e., $\eta = l\mu^* - Z$.

The traditional objective of the bandit problem is to maximize total reward given a specified number of pulls, $l$, to be made hence minimizing regret. [Lai & Robbins, 1985] showed that the regret should grow at least logarithmically and provided policies that attain the lower bound for specific probability distributions. [Agrawal, 1995] provided policies achieving the logarithmic bounds incorporating sample means that are computationally more efficient and [Auer et al., 2005] described policies that achieve the bounds uniformly over time rather than only asymptotically. Meanwhile, [Even-Dar et al., 2006] provided another quantification of the objective measuring quickness in determining the best arm. A Probably Approximately Correct (PAC) framework is incorporated quantifying the identification of an $\epsilon$-optimal arm with probability $1 - \delta$. The objective is then to minimize the sample complexity $l$, the number of samples required for such an arm identification. [Even-Dar et al., 2006] describe *Median Elimination Algorithm* achieving $O(n)$ sample complexity. This is further extended to finding $m$ arms that are $\epsilon$-optimal with high probability in [Kalyanakrishnan & Stone, 2010].

We propose a learning algorithm with $O(n)$ sample complexity that also achieves substantially low regrets during learning compared to the Median Elimination Algorithm in [Even-Dar et al., 2006]. We address the $n$-arm bandit problem with generic probability distri-butions without any restrictions on the means, variances or any higher moments that the distributions may possess. While [Even-Dar et al., 2006] provide a two parameter algorithm, we propose a single parametric algorithm that is based on fractional moments[1] of rewards acquired. To the best of our knowledge ours is the first work to use fractional moments in bandit problems, while recently we came to know that they have been used in the literature in other contexts. [Min & Chrysostomos, 1993] describe applications of such fractional (low order) moments in signal processing. It has been employed in many areas of signal processing including image processing [Achim et al., 2005] and communication systems [Xinyu & Nikias, 1996, Liu & Mendel, 2001]. Our algorithm using fractional moments is theoretically shown to have $O(n)$ sample complexity in finding an $\epsilon$-optimal arm. Experiments also show that the algorithm substantially reduces regret during learning. A brief overview of what is presented: Section 2 describes motivations for the proposed algorithm, followed by theoretical analysis of optimality and sample complexity in Sections 3 and 4. Finally, experimental results and observations are presented in Section 5 followed by conclusions and scope for future work.

## 2 PROPOSED ALGORITHM & MOTIVATION

Consider a bandit problem with $n$ arms with action $a_i$ denoting choice of pulling the $i^{th}$ arm. An experiment involves a finite number of arm-pulls in succession. In a particular such experiment, let $r_{i,k}$ be a sample of the reward acquired when $i^{th}$ arm was pulled for the $k^{th}$ time, $R_i$ being the associated random variable (with bounded support) for reward while taking action $a_i$. Then we have estimated means and variances for various actions as

$$\hat{\mu}_i = \frac{\sum_k r_{i,k}}{\sum_k 1} \;\; and \;\; \hat{\sigma}_i^{\,2} = \frac{\sum_k (r_{i,k} - \hat{\mu}_i)^2}{\sum_k 1}$$

When deciding about selecting action $a_i$ over any other action $a_j$, we are concerned about the rewards we would receive. Though $E(R_i)$ and $E(R_j)$ are indicative of rewards for the respective actions, variances $E[(R_i - \mu_i)^2]$ and $E[(R_j - \mu_j)^2]$ would give more information in the beginning stages of learning when the confidence in estimates of expectations would be low. In other words, we wouldn't have explored enough for the estimated expectations to reflect true means.

Though mean and variance together provide full knowledge of the stochasticity for some distributions, for instance Gaussian, we would want to handle generic distributions hence requiring to consider additional higher moments. It is the distribution after all, that gives rise to all the moments completely specifying the random variable. Thus we look at a generic probability distribution to model our exploration policy.

Consider selection of action $a_i$ over all other actions. For action $a_i$ to have a higher reward than any other action selected, the following probability holds:

$$\begin{aligned} M_i &= P\Big\{\bigcap_{j \neq i}(R_i > R_j)\Big\} \\ &= \prod_{j \neq i} P(R_i > R_j) \end{aligned} \quad (1)$$

where the independence of $R_j$'s is used. Thus, we can perform action selection based on the quantities $M_i$ provided we know the probability distributions, for which we propose the following discrete approximation: After selecting actions $a_i$, $a_j$ for $n_i$, $n_j$ times respectively, we can, in general, compute the probability estimate

$$\hat{P}(R_i > R_j) = \sum_{k \in N_i} \Big\{\hat{P}(R_i = r_{i,k}) \sum_{l \epsilon L_{i,k}^j} \hat{P}(R_j = r_{j,l})\Big\} \quad (2)$$

where sets $N_i$ and $L_{i,k}^j$ are given by,

$$N_i = \{k : 1 \leq k \leq n_i \text{ and } r_{i,k} \text{ are unique}\}$$

$$L_{i,k}^j = \{l : r_{j,l} < r_{i,k} \text{ and } r_{j,l} \text{ are unique}\}$$

with random estimates $\hat{P}(R_i = r_{i,k})$ of the true probability $P(R_i = r_{i,k})$ calculated by

$$\hat{P}(R_i = r_{i,k}) = \frac{|\{l : r_{i,l} = r_{i,k}\}|}{n_i} \quad (3)$$

Thus, with (1), (2) and (3) we can use $M_i$'s as preferences to choose action $a_i$'s.

Note that thus far we are only taking into account the probability, ignoring the magnitude of rewards. That is, if we have two instances of rewards, $r_{j,l}$ and $r_{m,n}$ with $P(R_j = r_{j,l}) = P(R_m = r_{m,n})$, then they contribute equally to the probabilities $P(R_i > R_j)$ and $P(R_i > R_m)$, though one of the rewards could be much larger than the other. For fair action selections, let us then formulate the preference function for action $a_i$ over action $a_j$ as,

$$A_{ij} = \sum_{k \in N_i} \Big\{\hat{P}(R_i = r_{i,k}) \sum_{l \epsilon L_{i,k}^j} (r_{i,k} - r_{j,l})^\beta \hat{P}(R_j = r_{j,l})\Big\}$$

---
[1]The $i^{th}$ moment of a random variable $R$ is defined as $E[R^i]$. Fractional moments occur when the exponent $i$ is fractional (rational or irrational).

**Algorithm 1** Policy Learning Algorithm

---
INITIALIZATION: Choose each arm once
DEFINE: $r_{i,k}$, the reward acquired for $k^{th}$ selection of arm $i$; the sets $N_i = \{k : r_{i,k}\ are\ unique\}$ and $L_{i,k}^j = \{l : r_{j,l} < r_{i,k}\ and\ r_{j,l}\ are\ unique\}$; and $n_i$, the number of selections made for arm $i$
LOOP:

1. $\hat{p}_{ik} = \hat{P}(R_i = r_{i,k}) = \frac{|\{l : r_{i,l} = r_{i,k}\}|}{n_i}$ for $1 \le i \le n$, $1 \le k \le n_i$

2. $A_{i,j} = \sum_{k \in N_i} \left\{ \hat{p}_{ik} \sum_{l \in L_{i,k}^j} (r_{i,k} - r_{j,l})^\beta \hat{p}_{jl} \right\}$ for $1 \le i, j \le n$, $i \ne j$

3. $A_i = \prod_{j \ne i} A_{i,j}\ \forall 1 \le i \le n$

4. Perform Probabilistic Action selection based on the quantities $A_i$

---

where $\beta$ determines how far we want to distinguish the preference functions with regard to magnitude of rewards. This way, for instance, arms constituting a higher variance are given more preference. For $\beta = 1$, we would have $A_{ij}$ proportional to,

$$E[R_i - R_j | R_i > R_j]$$

when the estimates $\hat{P}$ approach true probabilties. Hence, our preference function to choose $a_i$ over all other actions becomes,

$$A_i = \prod_{j \ne i} A_{ij}$$

The proposed Algorithm 1 thus picks arm $i$ with probability $A_i$ (normalized). We call this probabilistic action selection on the quantities $A_i$. To conclude, the algorithm is based on the set of rewards previously acquired but can be incrementally implemented (Results discussed in Section 5 use an incremental implementation). Also, the algorithm is invariant to the type of reward distributions. Besides incremental implementations, the algorithm and the associated computations simplify greatly when the underlying reward is known to follow a discrete probability distribution. Traditionally the motivation for using fractional moments is that they lead to more stable estimators. Fractional moments also seem to capture information from all integral moments, making them appropriate for usage in the proposed algorithm. We have strong reason to believe that this is the case, but have not been able to establish this conclusively.

The algorithm is shown to have a sample complexity of $O(n)$ in the Section 4. The theoretical bounds are arrived on a variant where the action selection is greedy with respect to quantities $A_i$. We believe the sample complexity bounds will further tighten for the probabilistic algorithm, which is also observed from the experiments detailed in Section 5.

## 3 OPTIMALITY

Consider a set of $n$ bandit arms with $R_i$ representing the random reward on pulling arm $i$. We assume that the reward is binary, $R_i \in \{0, r_i\} \forall i$. Following usual practice, we present the proof for the binary case only. The results extend to the case of general rewards with little additional work. Also, denote $p_i = P\{R_i = r_i\}$, $\mu_i = E[R_i] = r_i p_i$, and let us have $\mu_1 > \mu_2 > \cdots > \mu_n$ for simplicity.

### 3.1 CHERNOFF HOEFFDING BOUNDS

[Hoeffding, 1963] Let $X_1, X_2, \ldots X_n$ be random variables with common range $[0, 1]$ and such that $E[X_t | X_1, \ldots, X_{t-1}] = \mu$ for $1 \le t \le n$. Let $S_n = \frac{X_1 + \cdots + X_n}{n}$. Then for all $a \ge 0$ we have the following,

$$P\{S_n \ge \mu + a\} \le e^{-2a^2 n}$$
$$P\{S_n \le \mu - a\} \le e^{-2a^2 n}$$

### 3.2 $\epsilon$-OPTIMALITY OF THE ALGORITHM WITH PROBABILITY $1 - \delta$

Define,

$$I_{ij} = \begin{cases} 1 & if\ r_i > r_j \\ 0 & otherwise \end{cases}$$

$$\delta_{ij} = \begin{cases} \frac{r_j}{r_i} & if\ r_i > r_j \\ 1 & otherwise \end{cases}$$

We then have

$$\begin{aligned} A_{ij} &= I_{ij} \hat{p}_i \hat{p}_j (r_i - r_j)^\beta + \hat{p}_i (1 - \hat{p}_j)(r_i - 0)^\beta \\ &= \hat{p}_i \hat{p}_j (r_i - \delta_{ij} r_i)^\beta + \hat{p}_i (1 - \hat{p}_j) r_i^\beta \\ &= \hat{p}_i r_i^\beta \{1 + \hat{p}_j[(1 - \delta_{ij})^\beta - 1]\} \end{aligned}$$

The action selection function becomes,

$$A_i = (\hat{p}_i r_i^\beta)^{n-1} \prod_{j \ne i} \{1 + \hat{p}_j[(1 - \delta_{ij})^\beta - 1]\}$$

We need to bound the probability of the event $\{A_i > A_1 : i \in \{2, 3, \ldots, n\}\}$, since arm-1 is optimal. Let $i \ne 1$ be an arm which is not $\epsilon$-optimal, $\mu_i < \mu_1 - \epsilon$.

The policy would choose arm $i$ instead of the best arm 1, if we have

$$\begin{aligned}
A_i &> A_1 \\
(\hat{p}_i r_i^\beta)^{n-1} &> (\hat{p}_1 r_1^\beta)^{n-1} \frac{\prod_{j\neq 1}\{1+\hat{p}_j[(1-\delta_{1j})^\beta - 1]\}}{\prod_{j\neq i}\{1+\hat{p}_j[(1-\delta_{ij})^\beta - 1]\}} \\
&= (\hat{p}_1 r_1^\beta)^{n-1} \zeta
\end{aligned}$$

where

$$\zeta = \frac{\prod_{j\neq 1}\{1+\hat{p}_j[(1-\delta_{1j})^\beta - 1]\}}{\prod_{j\neq i}\{1+\hat{p}_j[(1-\delta_{ij})^\beta - 1]\}} \quad (4)$$

Note that for any $j$, $\delta_{ij} > \delta_{1j}$. Thus $\zeta > 1$ and using this, we get

$$\{(\hat{p}_i r_i^\beta)^{n-1} - (\hat{p}_1 r_1^\beta)^{n-1}\} = (\hat{p}_1 r_1^\beta)^{n-1}(\zeta - 1) > 0$$

Now, since $n-1 \in \mathbb{N}$, we have to bound the probability of the event $\hat{p}_i r_i^\beta > \hat{p}_1 r_1^\beta$.

### 3.2.1 Case $\beta = 1$

With $\beta = 1$, we are required to bound the probability of the event $\hat{p}_i r_i > \hat{p}_1 r_1$, which is same as the event $\hat{\mu}_i > \hat{\mu}_1$. Thus, the probability of choosing an arm $i$ that is not $\epsilon$-optimal is,

$$\begin{aligned}
P\{\hat{\mu}_i > \hat{\mu}_1\} &\leq P\{\hat{\mu}_i > \mu_i + \epsilon/2 \text{ or } \hat{\mu}_1 < \mu_1 - \epsilon/2\} \\
&\leq P\{\hat{\mu}_i > \mu_i + \epsilon/2\} \\
&\quad + P\{\hat{\mu}_1 < \mu_1 - \epsilon/2\} \\
&\leq 2e^{-2(\epsilon/2)^2 l}
\end{aligned}$$

if we sample each arm $l$ times before the determination of an $\epsilon$-optimal arm. The last step uses Chernoff-Hoeffding bound. Now, choosing $l = (2/\epsilon^2)\ln(2n/\delta)$ ensures $P\{A_i > A_1\} \leq \frac{\delta}{n}$. Hence we have,

$$P\{A_j > A_1 : j \neq 1\} \leq \delta$$

Sample complexity corresponding to all arms is then $nl = O(n \ln n)$.

### 3.2.2 Case $\beta \neq 1$

With $\beta \neq 1$, we are required to bound the probability of the event $\hat{p}_i r_i^\beta > \hat{p}_1 r_1^\beta$, which is same as the event $\hat{\mu}_i r_i^{\beta-1} > \hat{\mu}_1 r_1^{\beta-1}$. Thus we need to bound the probability of the event $\hat{\mu}_i > \hat{\mu}_1(1+\gamma_i)$ where $\gamma_i$ is fixed and is $(r_1/r_i)^{\beta-1} - 1$ for a given $\beta$ that is varying around 1. Note that $-1 < \gamma_i < \infty$ and $\gamma_i(\beta-1) > 0$. Define $R_{1i} = R_1(1+\gamma_i)$, $\alpha_i = \mu_1 \gamma_i$ then $\mu_{1i} = \mu_1 + \alpha_i$. The values that $\beta$ can take is constrained by $\alpha_i > -\epsilon$. Then, we have

$$\mu_i < \mu_{1i} - (\epsilon + \alpha_i)$$

Since $\hat{\mu}_1(1+\gamma_i) = \hat{\mu}_{1i}$, the probability of choosing an arm $i$ that is not $\epsilon$-optimal is,

$$\begin{aligned}
P\{\hat{\mu}_i > \hat{\mu}_{1i}\} &\leq P\{\hat{\mu}_i > \mu_i + \frac{\epsilon+\alpha_i}{2} \text{ or} \\
&\quad \hat{\mu}_1(1+\gamma_i) < \mu_1(1+\gamma_i) - \frac{\epsilon+\alpha_i}{2}\} \\
&\leq P\{\hat{\mu}_i > \mu_i + \frac{\epsilon+\alpha_i}{2}\} + \\
&\quad P\{\hat{\mu}_1 < \mu_1 - \frac{\epsilon+\alpha_i}{2(1+\gamma_i)}\} \\
&\leq e^{-2(\frac{\epsilon+\alpha_i}{2})^2 l} + e^{-2(\frac{\epsilon+\alpha_i}{2(1+\gamma_i)})^2 l} \quad (5)
\end{aligned}$$

if we sample each arm $l$ times before the determination of an $\epsilon$-optimal arm. Now, choosing

$$l_i = \begin{cases} (2(1+\gamma_i)^2/(\epsilon+\mu_1\gamma_i)^2)\ln(2n/\delta) & if\ \gamma_i > 0 \\ (2/(\epsilon+\mu_1\gamma_i)^2)\ln(2n/\delta) & if\ \gamma_i < 0 \end{cases}$$

ensures $P\{A_i > A_1\} \leq \frac{\delta}{n}$. For $\gamma_i > 0$ and $\gamma_i < 0$, the second and first terms of equation (5) dominate respectively and we choose $l$ accordingly. Hence,

$$P\{A_j > A_1 : j \neq 1\} \leq \delta$$

Notice the co-efficients arrived at,

$$\begin{cases} \frac{2(1+\gamma_i)^2}{(\epsilon+\mu_1\gamma_i)^2}, & when\ \gamma_i > 0 \\ \frac{2}{(\epsilon+\mu_1\gamma_i)^2}, & when\ \gamma_i < 0 \end{cases}$$

The coefficient when $\gamma_i < 0$ dominates $2/\epsilon^2$ while the one when $\gamma_i > 0$ does not in the respective regions of applicability. That is, we see that the bound gets better or worse accordingly whether $\beta > 1$ or $\beta < 1$, keeping in mind that $\beta$ and in turn $\gamma_i$ has a limited range to vary upon. Consistent with this inference, we also note that when $\gamma_i < 0$, $\mu_{1i} < \mu_1$ and hence we have higher probability of not selecting the best action, arm 1. This means, for values of $\beta < 1$, the algorithm is actually trying to explore more. More generally, the algorithm explores or exploits progressively more as the value of $\beta$ is decreased or increased relatively.

## 4 SAMPLE COMPLEXITY

### 4.1 MODIFIED CHERNOFF HOEFFDING BOUNDS

Chernoff Hoeffding bounds on a set of dependent random variables is discussed for analysis of the proposed algorithm.

**Chromatic and Fractional Chromatic Number of Graphs** A proper vertex coloring of a graph $G(V, E)$ is assignment of colors to vertices such that no two vertices connected by an edge have the same color. Chromatic number $\chi(G)$ is defined to be the minimum number of distinct colors required for proper coloring of a graph.

A $k$-fold proper vertex coloring of a graph is assignment of $k$ distinct colors to each vertex such that no two vertices connected by an edge share any of their assigned colors. $k$-fold Chromatic number $\chi_k(G)$ is defined to be the minimum number of distinct colors required for $k$-fold proper coloring for a graph. The Fractional Chromatic number is then defined as,

$$\chi'(G) = \underset{k\to\infty}{Lt} \frac{\chi_k(G)}{k}$$

Clearly, $\chi'(G) \leq \chi(G)$

**Chernoff Hoeffding Bounds with Dependence** [Dubhashi & Panconesi, 2009] Let $X_1, X_2, \ldots X_n$ be random variables, some of which are independent, have common range $[0, 1]$ and mean $\mu$. Define $S_n = \frac{X_1 + \cdots + X_n}{n}$ and a graph $G(V, E)$ with $V = \{1, 2, \ldots, n\}$. Edges connect vertices $i$ and $j$, $i \neq j$ if and only if $X_i$ is dependent on $X_j$. Let $\chi'(G)$ denote the fractional chromatic number of graph $G$. Then for all $a \geq 0$ we have the following,

$$P\{S_n \geq \mu + a\} \leq e^{-2a^2 n/\chi'(G)}$$
$$P\{S_n \leq \mu - a\} \leq e^{-2a^2 n/\chi'(G)}$$

**Bounds on the proposed formulation** Consider $n$ arms where $k^{th}$ arm is sampled for $m_k$ times. Let $X_u^{(k)}$ be a random variable related to the reward acquired on the $u^{th}$ pull of $k^{th}$ arm. Then, for $\{m_k : 1 \leq k \leq n\}$, $X_1^{(k)}, X_2^{(k)}, \ldots X_{m_k}^{(k)}$ are random variables (with common range within $[0, 1]$, say) such that $E[X_t^{(k)} | X_1^{(k)}, \ldots, X_{t-1}^{(k)}] = \mu^{(k)} \forall k, t : 1 \leq t \leq m_k$ and $X_u^{(i)}$ is independent of $X_v^{(j)}$ for $i \neq j$. Define random variables $S_{m_k}^{(k)} = \frac{X_1^{(k)} + \cdots + X_{m_k}^{(k)}}{m_k}$, $T = \prod_k S_{m_k}^{(k)}$ and $U = \prod_k X_{i_k}^{(k)}$ where $1 \leq i_k \leq m_k$. Let $U_1, U_2, \ldots, U_q$ be the realizations of $U$ for different permutations of $i_k$ in the respective ranges where $q = \prod_k m_k$. Then we have

$$T = \frac{\sum_1^q U_l}{q}$$

Also, $\mu_T = E[T] = \prod_k \mu^{(k)}$. Construct a graph $G(V, E)$ on $V = \{U_i : 1 \leq i \leq q\}$ with edges connecting every pair of dependent vertices. Let $\chi(G)$ and $\chi'(G)$ denote the chromatic and fractional chromatic numbers of graph $G$ respectively. Applying Chernoff-Hoeffding bounds on $T$, we have for all $a \geq 0$,

$$P\{T \geq \mu_T + a\} \leq e^{\frac{-2a^2 q}{\chi'(G)}} \leq e^{\frac{-2a^2 q}{\chi(G)}}$$
$$P\{T \leq \mu_T - a\} \leq e^{\frac{-2a^2 q}{\chi'(G)}} \leq e^{\frac{-2a^2 q}{\chi(G)}}$$

## 4.2 TIGHTER BOUNDS ON SAMPLE COMPLEXITY

We need to bound the probability of the event

$$A_i > A_1$$

where $A_i = (\hat{p}_i r_i^\beta)^{n-1} \prod_{j \neq i} \{1 + \hat{p}_j[(1 - \delta_{ij})^\beta - 1]\}$. Define $B_{ij} = \{1 + \hat{p}_j[(1-\delta_{ij})^\beta - 1]\}$ and $C_i = (\hat{p}_i r_i^\beta)^{n-1}$.

The event under concern then is,

$$C_i \prod_{j \neq i} B_{ij} > C_1 \prod_{j \neq 1} B_{1j}$$

$C_i, B_{ij}$ for $1 \leq j \neq i \leq n$ are estimates of different random variables that are independent of one another. Let arm $k$ be chosen $m_k$ times, so $C_i, B_{ij}$ have $m_i$ and $m_j$ realizations respectively.

Assuming arms $i$ and $1$ are chosen for equal number of times, we define the random variable

$$T = C_i \prod_{j \neq i} B_{ij} - C_1 \prod_{j \neq 1} B_{1j}$$

which would have $q = \prod_j m_j$ realizations and true mean $\mu_T = E[T] = E[A_i] - E[A_1]$. Further, as seen in subsection 4.1, $T$ can be decomposed as

$$T = \frac{\sum_1^q U_l}{q}$$

with $U_l$ being a product of $n$ independent random variables $X_{i_k}^{(k)}, 1 \leq k \leq n$ for different permutations of $i_k$.

The event to be bounded in probability is then

$$T \geq 0$$

Using the modified Chernoff-Hoeffding Bounds,

$$P(T \geq 0) = P(T \geq \mu_T - \mu_T) \leq e^{-2\mu_T^2 q/\chi(G)} \quad (6)$$

where the graph $G(V, E)$ is constructed on $V = \{U_i : 1 \leq i \leq q\}$ with edges connecting every pair of dependent vertices. While $|V| = q$, we see that each vertex is connected to $q - \prod_k(m_k - 1)$ other vertices. Then, the total number of edges in the graph becomes, $|E| = \frac{1}{2} n \left(q - \prod_k(m_k - 1)\right)$.

Table 1: Reduction in Complexity achieved

| $n$ | Weaker Bounds: $n(\ln n)$ | Proposed Algorithm: $n(n\ln^2 n)^{1/n}$ |
|---|---|---|
| 1,000 | 7,000 | 1,010 |
| 1,000,000 | 14,000,000 | 1,000,010 |

Consider the number of ways a combination of two colors can be picked, $^{\chi(G)}C_2 = \frac{1}{2}\chi(G)(\chi(G)-1)$. For optimal coloring, there must exist at least one edge connecting any such pair of colors picked. So,

$$\frac{1}{2}\chi(G)(\chi(G)-1) \leq |E|$$
$$\chi(G)(\chi(G)-1) \leq n\Big(q - \prod_k(m_k - 1)\Big)$$

Since $(\chi(G)-1)^2 \leq \chi(G)(\chi(G)-1)$, we have

$$(\chi(G)-1)^2 \leq n\Big(q - \prod_k(m_k - 1)\Big)$$
$$\chi(G) \leq 1 + \sqrt{n\Big(q - \prod_k(m_k - 1)\Big)}$$

Hence from (6),

$$P(T \geq 0) \leq exp\left(\frac{-2\mu_T^2 q}{1 + \sqrt{n(q - \prod_k(m_k - 1))}}\right)$$

In a simple case when each arm is sampled $l$ times, $m_k = l\ \forall k$ and

$$P(T \geq 0) \leq exp\left(\frac{-2\mu_T^2 l^n}{1 + \sqrt{n(l^n - (l-1)^n)}}\right)$$
$$\leq exp\left(\frac{-2\mu_T^2 l^n}{2\sqrt{nl^n}}\right)$$
$$\leq exp\left(\frac{-\mu_T^2 l^{n/2}}{\sqrt{n}}\right)$$

To get the sample complexity, sample each arm so that

$$exp\left(\frac{-\mu_T^2 l^{n/2}}{\sqrt{n}}\right) = \frac{\delta}{n}$$
$$\implies l = \left(\frac{n}{\mu_T^4}\ln^2\left(\frac{n}{\delta}\right)\right)^{\frac{1}{n}} \quad (7)$$

Sample complexity corresponding to all arms is then $nl$. The parameter $\epsilon$ is captured within $\mu_T$. To have a feel for the reduction in complexity achieved, we have Table 1.

Thus we see the complexity very similar to $O(n)$, which we prove below. We can improve the complexity further through successive elimination method discussed

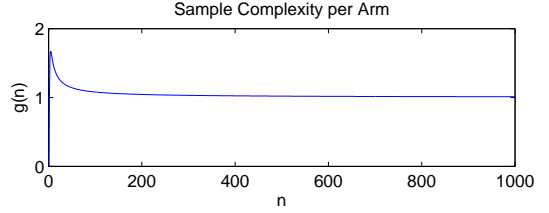

Figure 1: $g(n)$ versus $n$

in [Even-Dar et al., 2006]. But instead, consider

$$g(n) = (n\ln^2 n)^{\frac{1}{n}}$$

$$\implies \ln g(n) = \frac{\ln(n\ln^2 n)}{n} = \frac{\ln n + 2\ln\ln n}{n}$$
$$\frac{1}{g(n)}g'(n) = \frac{1 + \frac{2}{\ln n} - (\ln n + 2\ln\ln n)}{n^2}$$

Thus, $g$ is a decreasing function for $n > n_0 \in \mathbb{N}$ with the limit at infinity governed by

$$\underset{n\to\infty}{Lt}\ln(g(n)) = \underset{n\to\infty}{Lt}\frac{\ln n + 2\ln\ln n}{n}$$
$$= \underset{n\to\infty}{Lt}\frac{\frac{\partial}{\partial n}(\ln n + 2\ln\ln n)}{\frac{\partial}{\partial n}n}$$
$$= \underset{n\to\infty}{Lt}\left(\frac{1}{n} + \frac{2}{n\ln n}\right)$$
$$= 0$$
$$\implies \underset{n\to\infty}{Lt}g(n) = 1$$

With some numerical analysis, we see that $\forall n > 5$, $g'(n) < 0$. Also, $g$ attains a maximum at $n = 5$, with $g(n) = 1.669$. A plot of $g$ versus $n$ is shown in Figure 1. So $g(n) < 1.67\ \forall n \in \mathbb{N}$. Thus we have

$$O(n(n\ln^2 n)^{\frac{1}{n}}) = O(\rho n)$$

where $\rho = 1.67$. Thus, Sample Complexity of the proposed algorithm is essentially $O(n)$.

## 5 EXPERIMENT & RESULTS

A 10-arm bandit test bed with rewards formulated as Gaussian with varying means and a variance of 1 was developed for the experiments. Probabilistic and Greedy action selections were performed on the quantities $A_i$. Figure 2 shows Average rewards and Cumulative Optimal Selections with plays averaged over 2000 tasks on the test bed. $\beta = 0.85$ was empirically chosen without complete parameter optimization (though 4 trials of different $\beta$ were made). The temperature of the SoftMax algorithm, $\tau = 0.24$ was observed to be best among the 13 different temperatures that were attempted for parameter-optimization of the SoftMax procedure. This temperature value was also seen to better $\epsilon$-greedy action selection with $\epsilon = 0.1$.

To see a more asymptotic performance, the number of plays was further increased to 5000 with Gaussian rewards incorporating varying means as well as variances and the corresponding plots are shown in Figure 3. As can be seen, though the proposed algorithm could not keep up in optimal selections, but yet is reaping higher cumulative rewards even 3000 turns after $\epsilon$-greedy finds better optimal selections (at around $1500^{th}$ turn). Also, comparisons with respect to the Median Elimination Algorithm [Even-Dar et al., 2006] which also achieves $O(n)$ sample complexity is shown in Figure 3. The parameters for the Median Elimination algorithm depicted ($\epsilon = 0.95, \delta = 0.95$) performed best among 34 different uniformly changing instances tested. Thus, we conclude that the proposed algorithm, in addition to incurring a low sample complexity in finding an $\epsilon$-optimal arm, substantially reduces regret while learning.

# 6 CONCLUSIONS & FUTURE WORK

The proposed algorithm, the first to use fractional moments in bandit literature to the best of our knowledge, provides PAC guarantees with $O(n)$ complexity in finding an $\epsilon$-optimal arm. Experimental results show the algorithm achieves substantially lower regrets not only when compared with parameter-optimized $\epsilon$-greedy and SoftMax methods but also with other $O(n)$ algorithms like [Even-Dar et al., 2006]. Minimizing regret has been a crucial factor in various applications. For instance [Agarwal et al., 2008] describe relevance to content publishing systems that select articles to serve hundreds of millions of user visits per day. Thus, finding regret bounds on the proposed algorithm would theoretically aid in the analysis of high cumulative rewards observed in the simulations.

We note that as the reward distributions are relaxed from $R \in \{0, r_i\}$ to continuous probability distributions, the number of terms in the numerator as well as the denominator products of $\zeta$ in (4) would increase quadratically and lead to an increase in $\zeta$. These changes would reflect in $\mu_T$ and thereby improve the sample complexity in (7). So we expect even lower sample complexities (in terms of the constants involved) with continuous reward distributions. But the trade off is really in the computations involved. The algorithm presented can be implemented incrementally, but requires that the set of rewards observed till then be stored. On the other hand the algorithm simplifies computationally to a much faster approach in case of

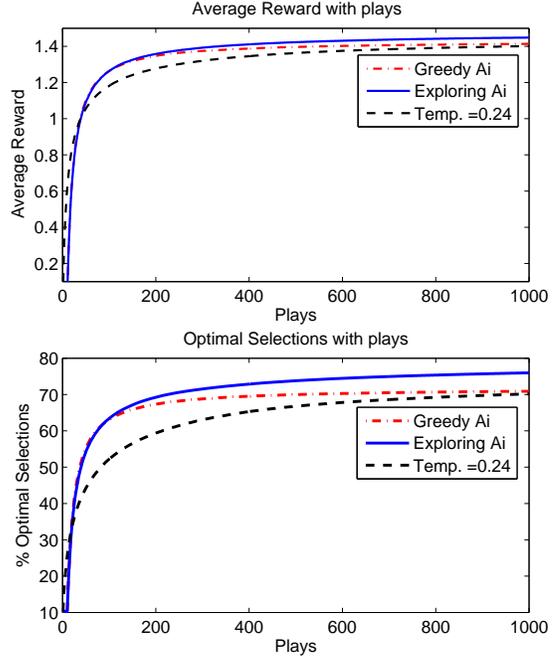

Figure 2: Two variants of the proposed algorithm, Greedy and Probabilistic action selections on $A_i$, are compared against the SoftMax algorithm

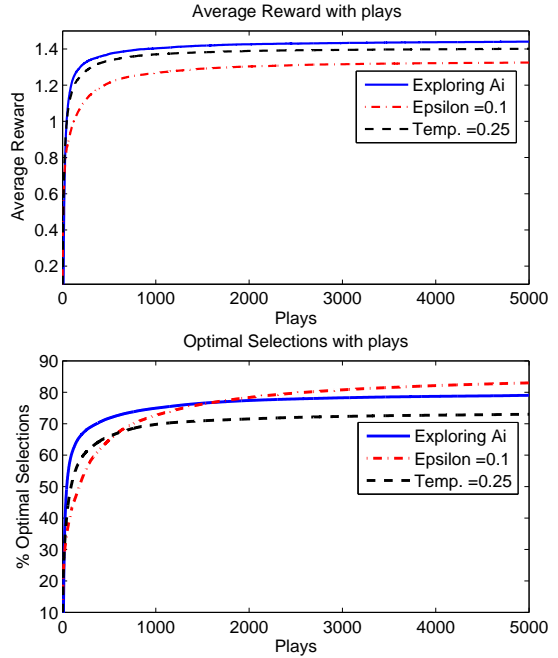

Figure 3: Asymptotic performance showing low Regret

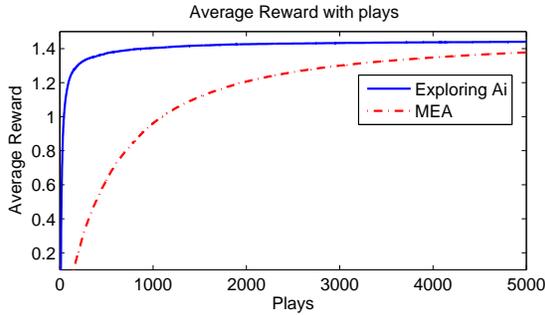

Figure 4: Comparison of the proposed algorithm with Median Elimination Algorithm (MEA) [Even-Dar et al., 2006]

Bernoulli distributions[2], as the mean estimates can be used directly. As the cardinality of the reward set increases, so will the complexity in computation figure. Since most rewards are encoded manually specific to the problem at hand, we expect low cardinal reward supports where the low sample complexity achieved would greatly help without major increase in computations.

We also note that the theoretical analysis assumes greedy action selections on the quantities $A_i$ and hence any exploration performed is unaccounted. Tighter theoretical bounds on the sample complexity could be arrived at by incorporating the exploration. Experiments have also shown that probabilistic action selections on $A_i$ prove better. Thus, theoretical bounds on the exploratory algorithm is another analysis to pursue.

---

[2]Essential requirement is rather a low cardinal Reward support